# Biomechanics-aware Multi-view Markerless Motion Capture of Dexterous Hand Movements

Pouyan Firouzabadi[1,2], J.D. Peiffer[1,2], Kunal Shah[2], Anton Sobinov[3], Lee E. Miller[1,2], R. James Cotton[2,4], Wendy M. Murray[1,2]

***Abstract***— Markerless motion capture (MMC) techniques have been widely beneficial in biomechanical analysis of human movement; however, application to complex motions of the hand lags other musculoskeletal systems. ***Objective:*** The primary goal of this study was to evaluate the performance of a biomechanical reconstruction method that implements a gradient-based optimization approach with a biomechanical model in the loop for tracking dexterous, unconstrained hand movements using MMC. ***Methods:*** Using a custom, 8-camera setup, we acquired 121 video recordings from 6 participants performing 11 different tasks that spanned 6 hand postures, 5 object manipulation tasks, and involved motion of the proximal upper limb joints. Performance of the proposed MMC pipeline was directly compared to a more commonly adopted two-stage reconstruction method that first triangulates 2D keypoints from computer vision pose estimation algorithms to 3D and then enforces biomechanical constraints by solving a constrained inverse kinematics problem. Relative performance was assessed qualitatively by visual inspection and quantitatively using a computer vision metric. ***Results:*** Our method generated solutions for all 121 video recordings; the two-stage method did not converge for 15% of the recordings. Across the remaining videos, our method produced more biomechanically plausible hand kinematics than the two-stage method and was more robust to occlusion effects during tasks that involved objects. ***Conclusion:*** The relative robustness of the end-to-end method suggests that it is more effective in utilizing the available 2D digital keypoint information. ***Significance:*** Automatic and biomechanically meaningful tracking of hand kinematics during dexterous movements has the potential to support clinical evaluation, rehabilitation monitoring, and studies of human motor control.

## I. Introduction

MARKERLESS motion capture (MMC) offers a practical means of capturing movement without any equipment worn on the body [1]. Specifically, MMC uses computer vision and pose estimation to algorithmically label anatomical landmarks on 2D video data acquired during movement [2], [3], [4], [5], [6], [7]. MMC uses these digital labels, referred to as keypoints, in place of the physical markers used in traditional motion capture, which reduces setup and time burdens on participants and investigators. These benefits can facilitate data collection in large-scale population studies [8] and in-the-wild athletic settings [9]. Despite the increasing adoption of MMC in studies of human movement, its application to complex hand motion lags other musculoskeletal systems. The human hand plays a central role in nearly all aspects of daily life, from fine motor tasks like writing to powerful grasps required for lifting. Understanding biomechanics and neural control of hand movement is essential and MMC is a promising technology to advance the field.

For traditional marker-based methods, a marker's location is physically linked to a specific anatomical landmark throughout the movement. In contrast, digital MMC keypoints are identified independently in each 2D video frame of a recorded movement. Because of this, the anatomical landmarks of interest can be labeled inconsistently across frames, resulting in anatomically impossible errors which must be addressed. Most approaches to biomechanical reconstruction of MMC data mitigate these errors in a two-stage analysis, first triangulating 2D keypoints from multiple cameras to 3D [10], then enforcing biomechanical constraints [11], [12], [13], [14], [15] by solving a constrained inverse kinematics problem. Instead, one recent development imposes biomechanical constraints in a single-stage process by iteratively tuning the pose and scaling of a 3D biomechanical model [16], [17], [18] and solving for the 3D joint kinematic trajectories that best reproject in to the 2D camera planes. This method has been described as "end-to-end" because it optimizes movement trajectories and model scaling all at once through a differentiable biomechanical model.

Both the two-stage and end-to-end reconstruction methods have previously been reported to achieve performance metrics equivalent to marker-based systems [11], [12], [16], [19]. Using computer vision metrics that focus on agreeability between reconstructed hand postures and pose estimation keypoint data [20], we previously demonstrated end-to-end optimization outperformed the two-stage fitting procedure for tracking complex hand postures [18]. However, we are not aware of any work that directly compares joint kinematics computed by both methods. The relative performance between methods is especially important for the hand because the small size of individual segments of the phalanges amplifies the sensitivity to keypoint errors [21]. In addition, object

"This work was supported by Northwestern University (NIH R01 NS131953), the Restore Center P2C (NIH P2CHD101913), and the Research Accelerator Program of the Shirley Ryan AbilityLab".

P. F. correspondence e-mail: pouyan@u.northwestern.edu

1 Biomedical Engineering Department, Northwestern University, Chicago, IL, US

2 Shirley Ryan AbilityLab, Chicago, IL, US

3 Department of Organismal Biology and Anatomy, University of Chicago, Chicago, IL, US

4 Department of Physical Medicine and Rehabilitation, Northwestern University, Evanston, IL.

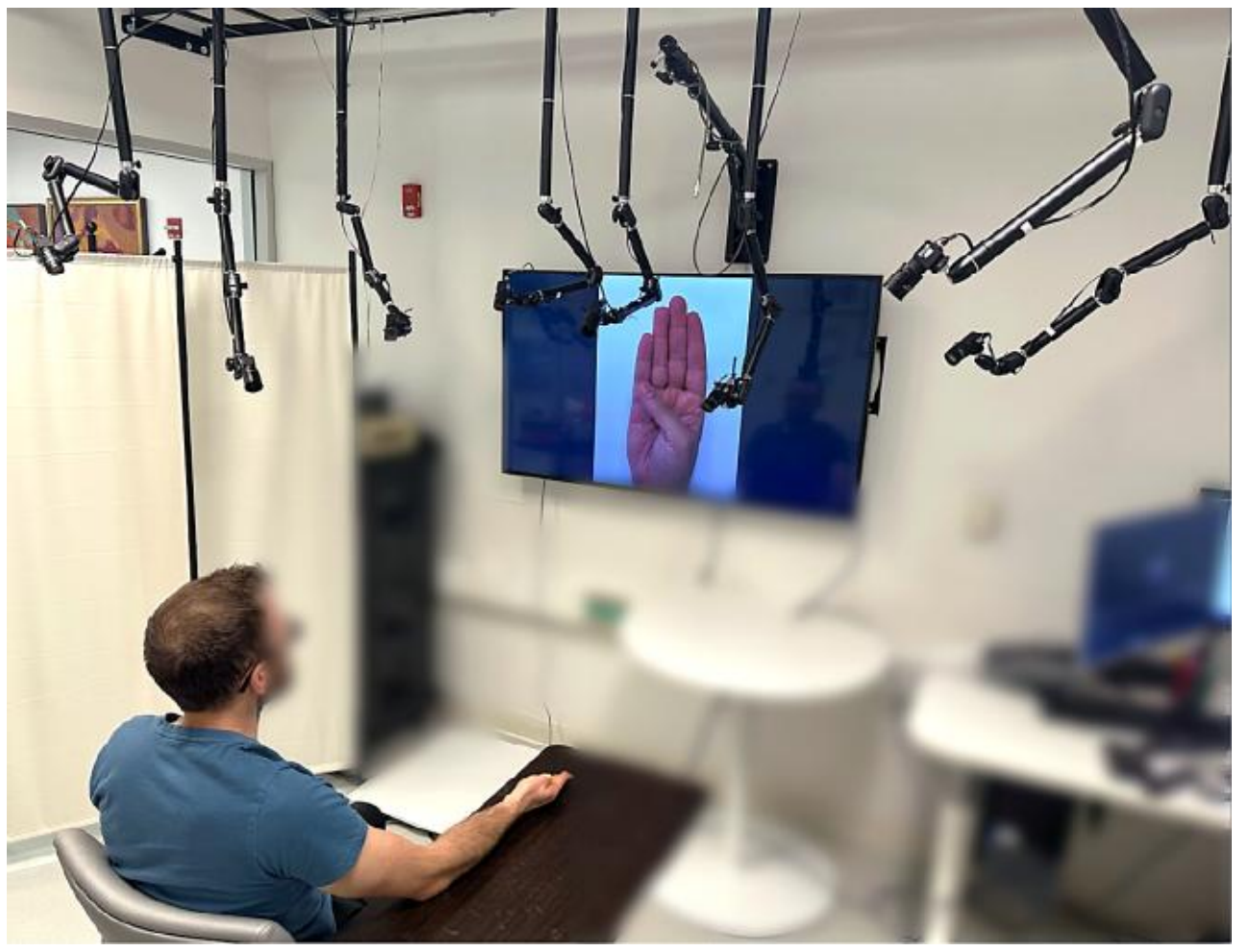

**Figure 1.** Experimental setup with 8 cameras surrounding the participant replicating the hand pose (the ASL letter B) shown on the video screen.

***Table 1.*** *Demographics of Healthy Participants*

| *Participant* | **Age** | **Sex** | **Arm** | **Skin Tone** [22] |
|---|---|---|---|---|
| ***P1*** | 26 | M | R | Tan |
| ***P2*** | 38 | M | R | Tan |
| ***P3*** | 25 | F | R | Light |
| ***P4*** | 25 | F | R | Light |
| ***P5*** | 23 | F | R | Dark |
| ***P6*** | 28 | M | R | Brown |

interaction – a critical aspect of hand function – introduces complexities to hand tracking. Only one previous study both incorporated biomechanical constraints into their MMC workflow and included object interaction [12]; however, the hand was constrained in a fixed upper limb posture while interacting with a single object. Comparison of existing reconstruction methods during functional upper limb motion and involving multiple, different objects is needed.

This work aims to evaluate the performance of two biomechanical reconstruction methods for tracking dexterous, unconstrained hand movements using MMC. Here, we demonstrate the feasibility of acquiring robust joint angles using the single-stage, end-to-end method across six participants performing both posture and object interaction tasks that include motion of more proximal joints of the upper limb. We directly compare kinematic performance to the more common two-stage method. Our work provides both quantitative and qualitative evidence that the end-to-end method more robustly tracks complex hand motions and interactions with objects.

## II. Method

This study received approval from the Northwestern University IRB board. Six able-bodied, right hand dominant participants with no history of injury or impairment to the arm or hand provided informed consent and completed the study (ref. Table 1).

### *A. Experimental Setup Protocol*

To collect videos of complex hand motions appropriate for biomechanical analyses, we used 8 FLIR BlackFly S GigE RGB cameras with a mixture of F1.4/6mm and F1.6/4.4-11mm variable focus lenses. This system records 2048x1536-pixel images at 30 frames per second, synchronized using the IEEE1558 protocol [23]. A 7x5 ChArucoBoard with 110 mm spacings was used for calibration [24]. Extrinsic and intrinsic matrices were estimated using the anipose library [25]. The calibrated camera parameters from multiple views allows for transformation between the 2D image planes and 3D space.

Participants were seated below the cameras, with their right arm resting on a table (Fig. 1). Camera lenses were set to capture the whole body with the right hand and upper arm in middle of the view. Participants were instructed to complete the following movements: form the American Sign Language (ASL) postures for letters A, B, D, F, L, and O, reach and grasp a tennis ball, a bottle, and a disk, and complete tasks linked to activities of daily living (key pinch and drinking). Participants formed ASL letters with the forearm fully supinated, rotated the forearm from supinated to fully pronated while maintaining the letter posture, and returned the forearm and hand back to supinated and neutral, respectively. For reach and grasp tasks, participants reached for the object, lifted it, and then released it on the table at its original position. For the key pinch task, participants were asked to use key pinch to grasp a key in a deadbolt mounted on the table, rotate it 180 degrees clockwise, rotate it back counterclockwise, and then release it. For the drinking task, participants reached for a cup, brought it to the mouth, and then set it back down on the table.

All participants completed every task in a single visit that involved multiple video recordings. A single video recording included the subject repeating one specific task three times in succession. Across all participants, 121 video recordings were collected. Two video recordings were collected for each individual ASL letter. The number of recordings for the remaining tasks varied between 1 and 2 among the different participants.

### *B. Keypoint Identification*

Videos were processed using PosePipe [26]. Since the hand movements we studied included proximal limb movements, the initial video processing used the MoVi keypoint set [27] available in the public whole-body pose estimation algorithm MeTRAbs-ACAE [28]. This analysis labeled the shoulder, clavicle, sternum, elbow, medial and lateral wrist, and metacarpophalangeal (MCP) joints. The hands of the study team were also visible in some camera views, so a bounding box that localized the right hand of the research participant was defined by extrapolating MoVi keypoints located at the wrist and MCP joints. Within this box, we further labeled 21 hand-specific keypoints using the RTMPose algorithm [3] from the MMPose library [29]. For each keypoint, RTMPose estimates the likelihood of the keypoint being present at that pixel location ranging from 0 to 1. For both reconstruction methods, we excluded any keypoint with likelihood $< 0.25$.

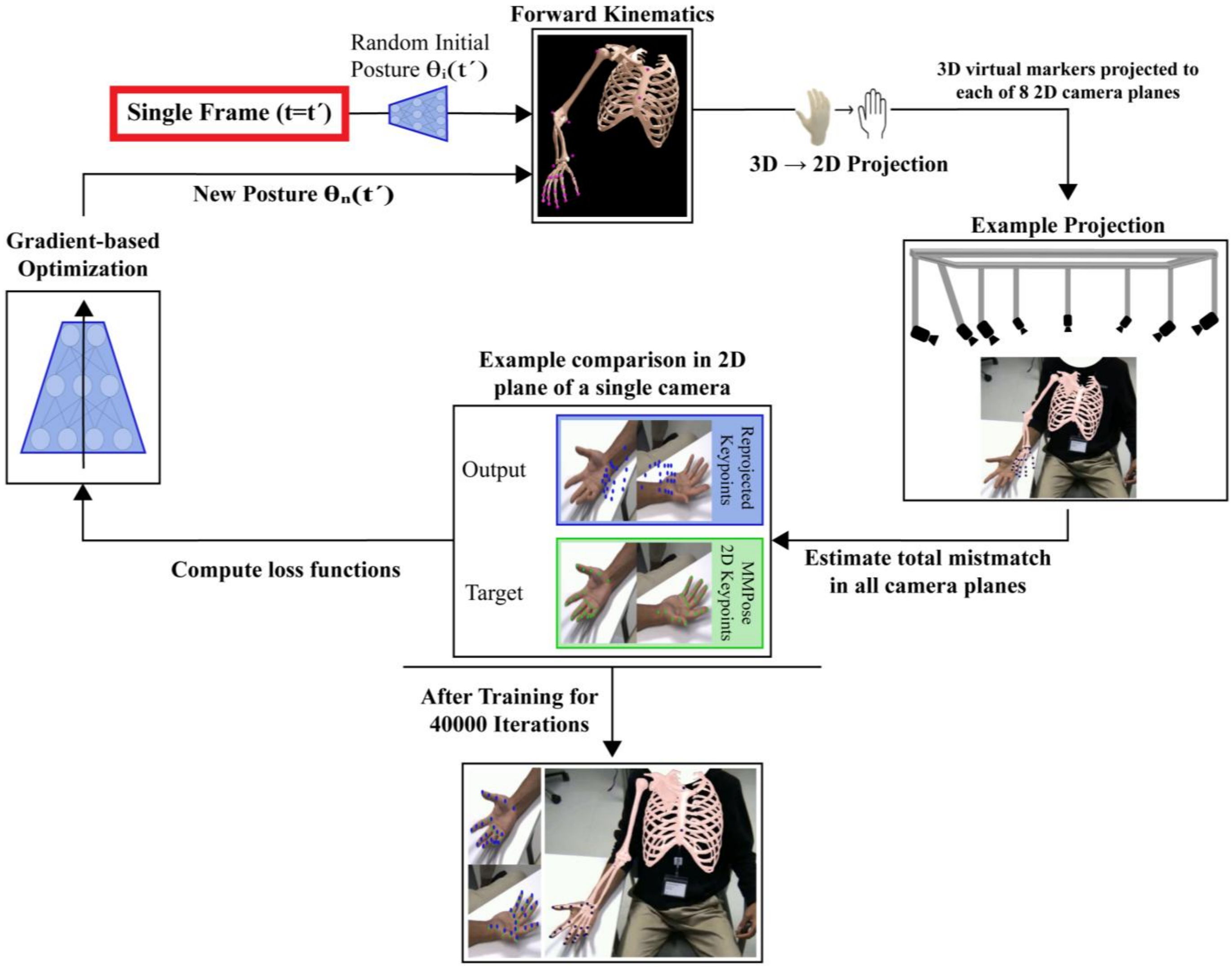

***Figure* 2.** Our end-to-end MMC reconstruction pipeline: For a single frame, the gradient-based optimizer, a multi-layer perceptron (MLP), initially outputs arbitrary joint angles to pose the biomechanical model. The model includes user-defined 3D virtual markers that correspond to the keypoints labeled via computer vision. When the model assumes the specified pose, the 3D coordinates of the virtual markers are reprojected to each of the 8 camera's 2D planes, allowing direct comparison to the 2D digital keypoints identified by the hand pose estimation algorithm for that frame. During the optimization process for a whole recording, the MLP learns a mapping from timestep (t) to joint angles $\theta(t)$, defining a best fit set of joint angle trajectories for the recording. In parallel with optimizing joint angles, the biomechanical model is scaled to the participant, including fine-tuning the location of the virtual markers on the model. The MLP is trained for 40000 iterations to overfit based on defined loss functions, producing final joint angles that best represent the 2D information available from all camera planes.

### *C. Arm and Hand Biomechanical Model and Simulation*

Both reconstruction methods implement the same biomechanical model with 28 user-defined virtual markers that correspond to the digital keypoints labeled by the pose estimation algorithms. Specifically, we implemented a 21 degree of freedom (DoF) model of the hand [13] combined with a 7-DoF model of the upper-extremity [14], both originally developed in OpenSim [32]. The combined model includes kinematic definitions of all finger and thumb movements (including coupled flexion for the carpometacarpal joints of the ring and little finger), wrist flexion/extension and deviation, forearm rotation (pronation/supination), elbow flexion, and shoulder DoFs that include the scapulohumeral rhythm. We include an additional 6 DoFs at the ground segment (the thorax) to represent its 3D position and orientation in the global frame.

### *D. Reconstruction Methods*

#### *1) End-to-end method*

Building on methods previously developed for whole body [17] and upper limb [19] tracking, we implemented a gradient-based optimization approach, including a Multi-Layer Perceptron (MLP), that integrates our biomechanical model in the loop to produce plausible 3D joint kinematic postures from the 8 camera views (Fig. 2). Initially, random joint angle guesses are generated by the MLP. These guesses specify the posture of the biomechanical model and the corresponding 3D coordinates of the user-defined virtual markers via forward kinematics. The 3D coordinates of the virtual markers are then reprojected to each camera's 2D plane. The error between the 2D reprojections and the 2D digital keypoints from the hand pose estimation algorithm is used to train the MLPs. Ultimately, the final trained MLP produces joint angles for the biomechanical model for each time step that best represents the 2D digital keypoints in all 8 camera views. During back propagation, a scaling matrix is computed which defines 6 isotropic scaling factors for the thorax, shoulder girdle, upper arm, forearm, hand, and fingers. In addition, the scaling matrix defines 28 offsets that adjust the positions of the virtual markers on the model. Model scaling and virtual marker offset adjustments are defined to best fit all video recordings of an individual participant using the bilevel approach developed by AddBiomechanics [33].

*a) Machine Learning Optimization Framework*

The gradient-based pipeline was implemented in Jax [34], using the Equinox framework [35]. We integrated the kinematic model (see II.C) with the machine learning framework in MuJoCo-MJX [36] after converting the model to MuJoCo [37] using the publicly available MyoConverter [38]. The MLP that produces the joint angles for forward kinematics and scales the model has seven layers of size (128, 256, 512, 1024, 2048, 2048, 4096; for more model architecture details refer to [17]).

For a single video recording (i.e., the recording of an individual participant repeating a specific task 3 times), the loss value to train the MLP included a reprojection loss function and two regularization terms (see [16], [17], [18] for details). Reprojection loss was defined as the average estimation error in pixels over all timesteps, cameras, and keypoints weighted by the confidence-level of the keypoints from the pose estimator. At each timestep (t) and for each camera (c), we calculate the estimation error for the kth-keypoint ($\delta_{t,k,c}$) as the Euclidean distance between the reprojected virtual marker ($\Pi_c x_{t,k}$) and labeled 2D keypoint ($y_{t,k,c}$) pixel coordinates as:

$$\delta_{t,k,c} = \left\| \Pi_c x_{t,k} - y_{t,k,c} \right\| \quad (1)$$

One of the regularization terms penalizes the use of joint kinematics that fall outside the joint constraints defined by the kinematic model. The second regularization term (an L2 loss function) limits the offset distance the 3D virtual markers can be shifted from their user-defined locations. The optimization was run in parallel for 40000 iterations for all video recordings from a specific participant's visit, to overfit the joint angles (for each recording) and find the best scaling factors and virtual marker offsets (for the whole session) [33] with the lowest loss value.

We performed optimization with the AdamW optimizer from Optax [39], using beta of 0.8 and weight decay of 1e-5. The learning rate included an exponential decay from an initial value of 1e-3 to an end value of 1e-6. The lambda for virtual marker offsets was set to 1e3. We performed this on an A6000 GPU and found it could perform more than 30 iterations per second over 20 recordings optimized jointly.

*2) Two-stage method*

We compared our end-to-end method to the more commonly adopted two-stage reconstruction method. We analyzed the same 2D information from the pose estimators using a robust triangulation technique [40] on all 8 cameras to transform 2D keypoints to 3D (Stage 1). The resulting 3D keypoint trajectories were then post-processed and served as inputs for a constrained inverse kinematics problem (Stage 2), solved using the OpenSim physics simulator [32] and the same biomechanical model and virtual markers described above (see Section II.C).

Digital keypoints triangulated to 3D in Stage 1 were smoothed by applying a median filter spanning 5 timepoints. Filtered 3D keypoints were used to generate trace files compatible with the OpenSim simulator. All settings and setup files used for scaling and optimization in OpenSim followed the publicly available OpenCap pipeline and previous work [11], [18]. Due to kinematic redundancy and keypoint noise, we observed convergence issues in Stage 2. To mitigate for this, we locked the initial pose of the ground segment of the model (thorax); then, the integrated inverse kinematics tool was used to extract joint angles as well as optimized 3D virtual marker locations reflecting the pose estimation keypoints.

*3) Performance metrics*

103 video recordings were used for direct comparisons between the two methods because 18/121 total recordings did not converge to a final solution for the two-stage method. Joint angle trajectories estimated using each method were compared quantitatively and qualitatively. First, we computed the absolute mean difference in joint angles between methods. To account for repeated and unequal numbers of measurements across tasks and participants, linear mixed-effects models were used. The model included joint as a categorical fixed effect and a subject-specific random intercept to account for within-population variability.

Because we did not have a gold standard to evaluate joint angle accuracy, we compared performance for an ASL posture (the letter B, demonstrated in Fig. 1) that required full extension of the fingers. We fit a linear mixed-effects model to evaluate the joint kinematics computed in the 10 trials of signing the letter B available from both reconstruction methods. The model included fixed effects for method (end-to-end vs. two-stage), joint level (distal IP, proximal IP, MCP), and their two-way interaction as well as a random intercept for participants.

We also used the percentage of correct keypoints (PCK), a computer vision metric [20], to evaluate the correspondence of reprojected 3D virtual marker locations to the 2D digital keypoints labeled by the pose estimator for each of the 103 video recordings. To limit the effects of inconsistencies in the performance of the human pose estimation algorithm, we only included those keypoints with a likelihood > 0.5 as determined by RTMPose. Specifically, for each of the 2D digital keypoints with likelihood > 0.5 identified across all timepoints and camera views for a given video recording (Total), we identified the number of reprojected virtual markers that fell within a 10-pixel threshold ($N_{true}$). That is,

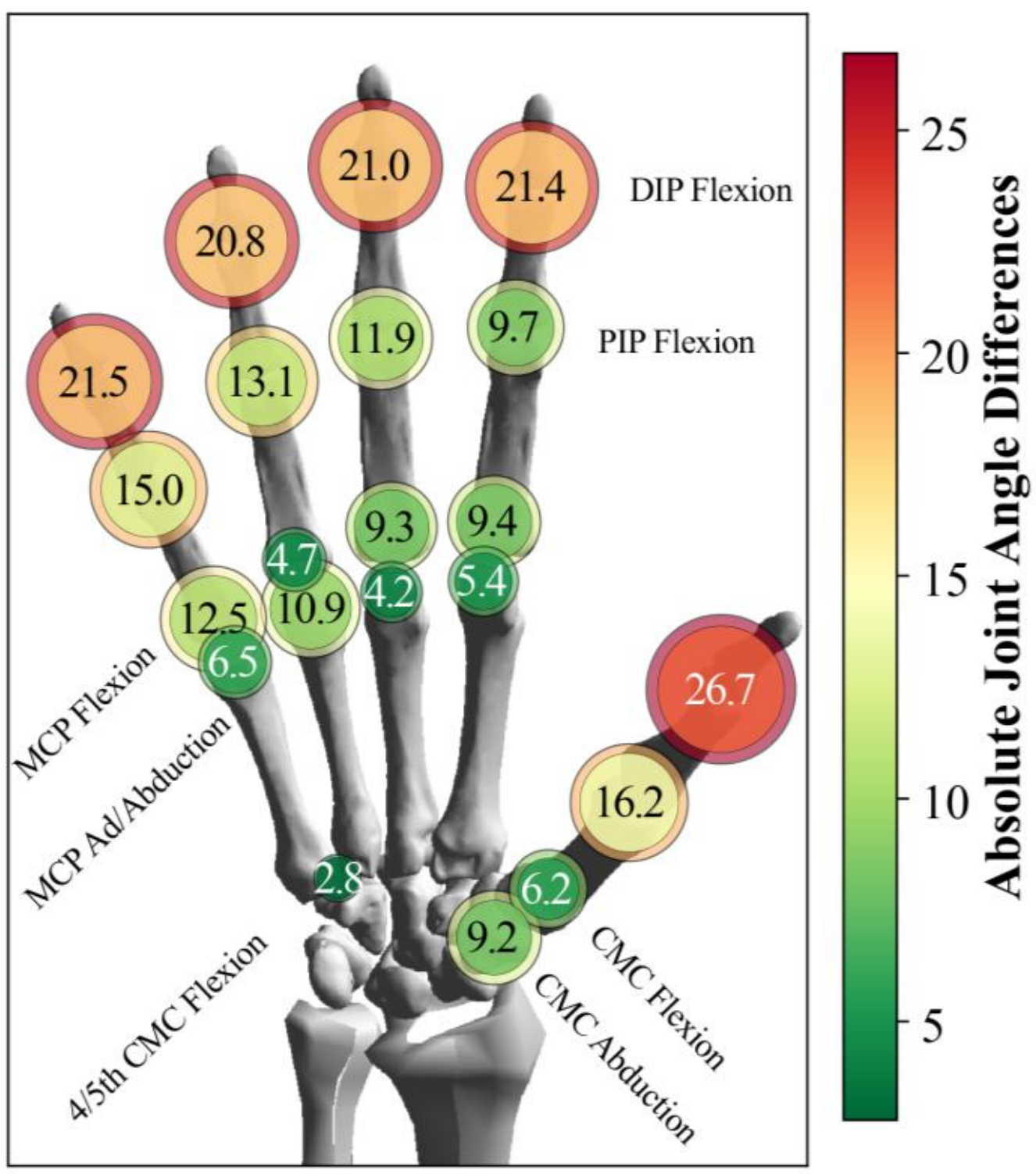

**Figure 3**. Absolute mean difference of individual joint angles produced by the end-to-end and two-stage reconstruction methods across all participants and recordings. The diameter and color of each inner circle reflects the magnitude of the average error. A separate ring, with diameter equal to the mean error + 1 standard deviation (s.d.) surrounds each filled circle. The color of the s.d. ring also indicates the value of the mean + 1 s.d. for that degree of freedom.

$$PCK = \frac{N_{\text{true}}}{\text{Total}} \times 100 \qquad (2)$$

Higher PCK values indicate better agreement. We tested whether tasks including objects affected PCK across reconstruction methods. Using a linear mixed-effects model we compared the statistical interaction effect between task paradigm (postures vs. objects) and analysis method (end-to-end vs. two-stage) while accounting for variability among participants as random intercept.

## III. Results

### *A. Joint Angles Comparison*

The two biomechanical reconstruction methods produced different joint angle estimates ($p < 0.001$). The average, absolute difference between the end-to-end and two-stage methods across all participants, recordings, timesteps, and joints was 12.1° ± 8.8°. The magnitude of the difference between methods depended on the joint. For example, we observed an average difference that ranged between 20° and 27° at the most distal joints, whereas differences between methods were less than 13° for both degrees of freedom (flexion/extension and ab/adduction) at the proximal base joints of each digit (Fig. 3).

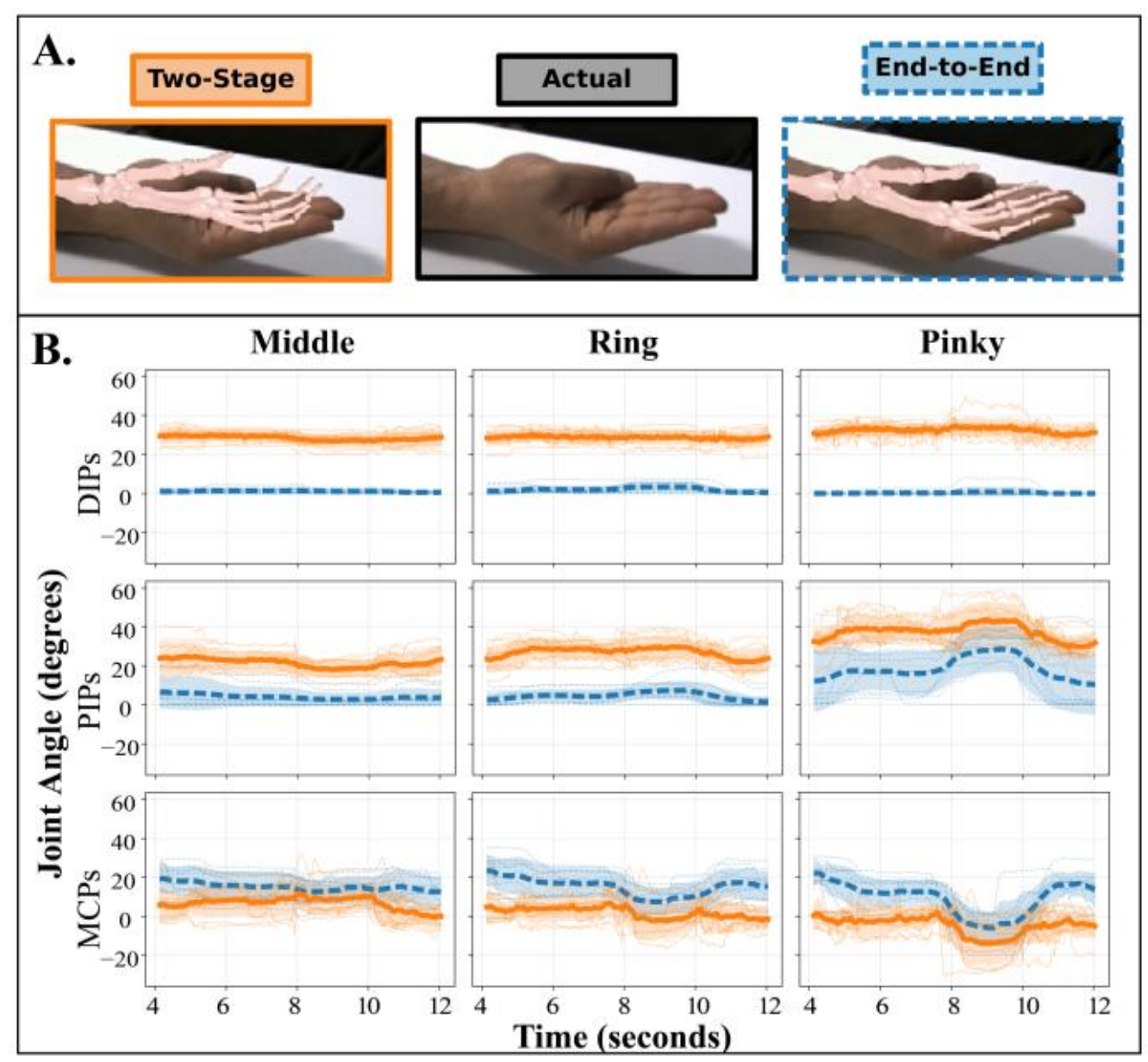

**Figure 4 A.** The biomechanical model, positioned as computed from both reconstruction methods and overlayed on the video of the participant forming the ASL posture for the letter B. All data is from the same video recording at t = 4s. For the two-stage method (solid orange outline), pronounced flexion is observed in more distal joints; the end-to-end result (dashed blue outline) for the same frame shows better agreement to the actual position of the hand. **B.** Computed joint angles for the middle, ring, and pinky fingers across 10 trials involving signing the letter B. Time (~4-12s) corresponds to the hand being maintained in the B posture during forearm pronation. Solid, bold, orange lines indicate the average computed via the two-stage method (shaded orange regions indicated one standard deviation, lighter lines show individual trials). Dashed, bold, blue lines indicate the average computed via the end-to-end method (shaded blue regions indicated one standard deviation, lighter lines show individual trials).

Visual inspection of the extension posture of the finger joints while forming the ASL letter B provides qualitative support that the end-to-end method better reflected the actual movement. For example, reprojection of the biomechanical model onto individual 2D video frames illustrates the two-stage method yields fingers with greater distal flexion than adopted by the participants (Fig. 4A). When quantified, the two-stage method exhibited significantly larger differences from full extension (0°) at the DIP joints for all four fingers (29.4º ± 2.4º; Fig. 4B shows data from 3 fingers) compared with the end-to-end (1º ± 1º; $p<0.001$). For the MCP joints, significant deviations from zero ($p < 0.001$) for the end-to-end method (e.g., average flexion between 4-12s = 13º ± 3º) better reflected hand posture than two-stage (3º ± 5.5º; cf., Fig. 4A, MCP joints are flexed for this participant).

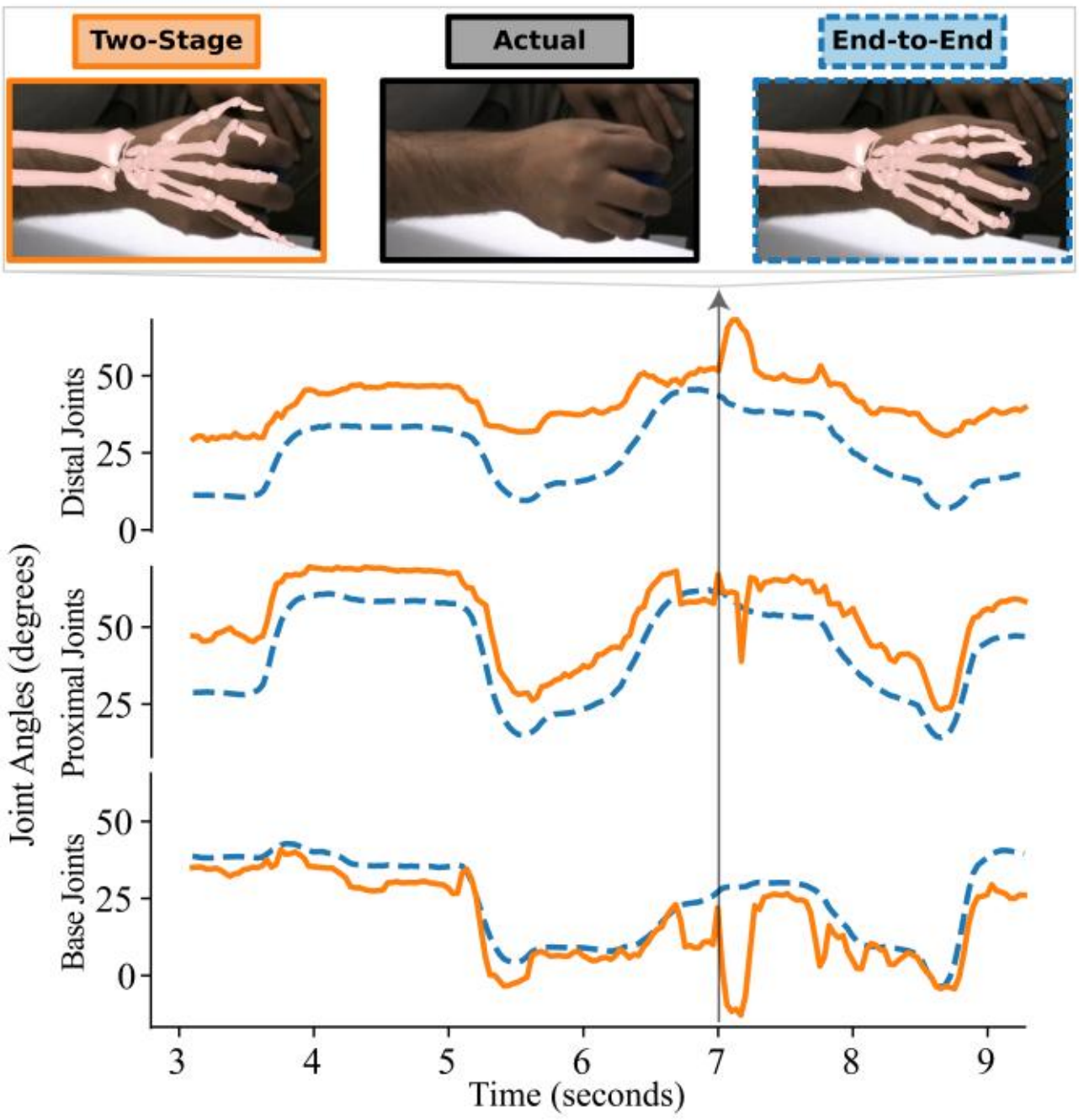

**Figure 5.** Summary of the joint angles that result from the two-stage (orange lines) and end-to-end (blue dashed lines) methods for a single object interaction task that involved reaching and grasping a ball. For both methods, the curves indicate joint angles averaged across groups of joints, including (top graph) the Distal joint angles (average of index, middle, ring, and pinky DIP joints + thumb IP joint), (middle graph) Proximal joint angles (average of all finger PIP joints plus thumb MCP joint), and (bottom graph) Base joint angles (all finger MCPs plus thumb CMC joint). Images at the top represent the hand posture at 7 seconds while holding the ball with the optimized biomechanical poses overlayed on each image.

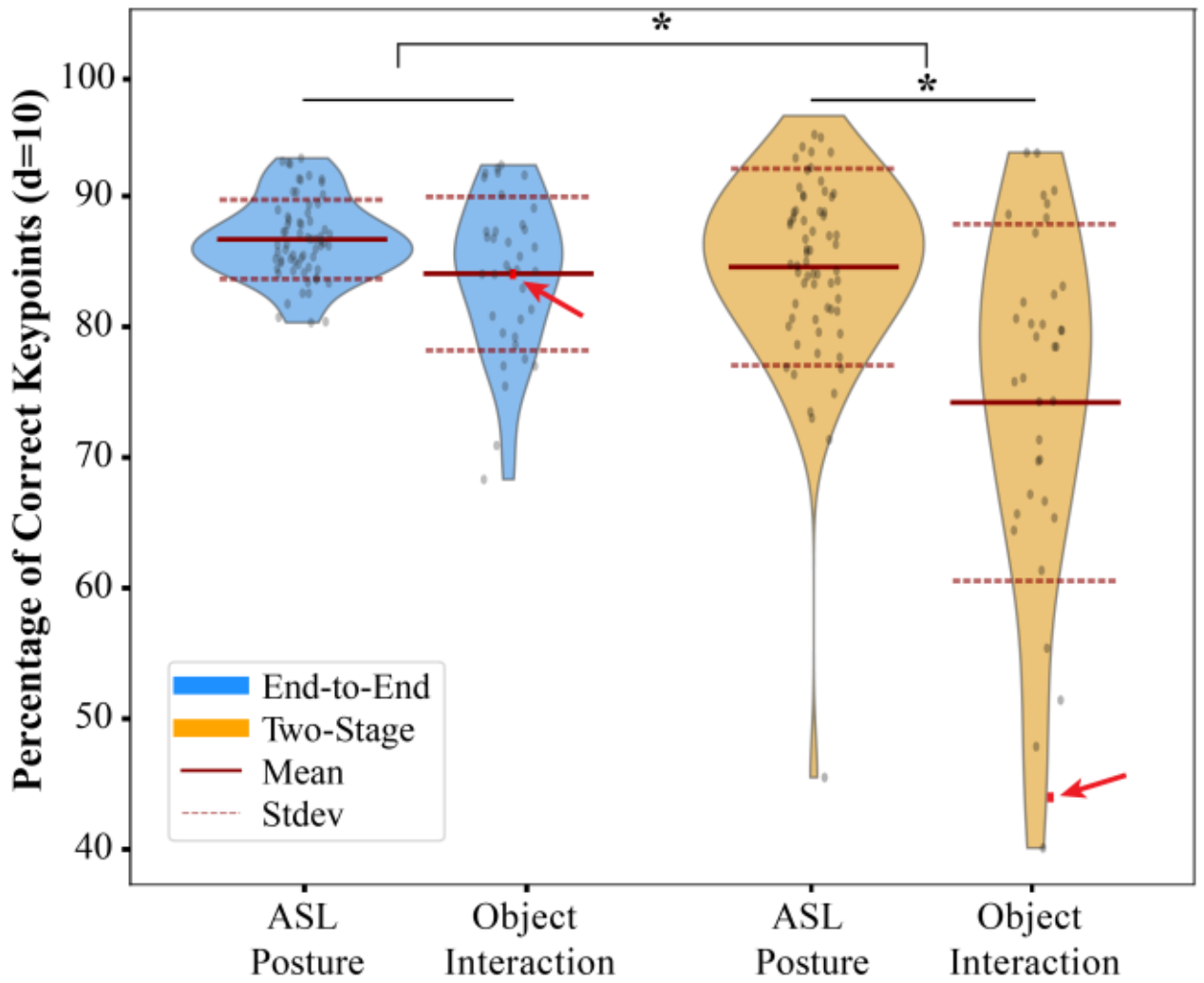

**Figure 6.** Percentage of correct keypoints (PCK) metric computed from the biomechanical reconstructions of the same 103 video recordings using the end-to-end (blue) and two-stage (orange) methods. PCK quantifies the percentage of reprojected 3D virtual markers that fall within a 10-pixel threshold of the 2D digital keypoints from pose estimation. Higher PCK values indicate better agreement. The two-stage method's performance on tasks that include objects significantly decreased relative to postures ($p < 0.001$). Red dots and arrows highlight the trial displayed in Figure 5.

### *B. Object interaction analysis*

During object interaction tasks, the two-stage method had considerably larger errors than our end-to-end method. Visual inspection of overlays of the joint kinematics from both reconstruction methods reveal the relative robustness of the end-to-end method to object occlusion effects (Fig 5). The pronounced deviation from the actual posture for the two-stage method is apparent, and much reduced by the end-to-end method. Across 103 video recordings acquired from 6 different participants performing 11 different tasks, the two-stage method's PCK score was significantly lower for tasks involving objects, and the statistical interaction effect (method x task paradigm) was significant ($p < 0.001$; Fig. 6).

## IV. Discussion

The primary goal of this study was to evaluate the performance of a recently developed, single-stage, end-to-end biomechanical reconstruction method for tracking dexterous, unconstrained hand movements using MMC. By integrating joint constraints directly into the optimization loop, this method eliminates the need to rely on post-processing (e.g., filtering) to reduce non-physiological errors in triangulated keypoints [12], [19]. Our results demonstrate that our end-to-end pipeline consistently outperformed a more traditional, two-stage method for both complex postures and object interaction tasks. First, our pipeline generated solutions for the full set of 121 video recordings. In contrast, Stage 2 of the two-stage method (the constrained inverse kinematics optimization) did not converge for 15% (18/121) of the recordings, despite applying commonly adopted post-processing techniques to the triangulated keypoints from Stage 1 (see Methods, Section D.2). Across the remaining 103 video recordings, our end-to-end method produced more biomechanically plausible hand kinematics than the two-stage method. This was particularly evident at the distal finger joints, where the two-stage method introduced excessive flexion for a posture where these joints were extended. The end-to-end approach was also more robust for object interaction tasks. Our analysis exposed a key limitation of the two-stage reconstruction method: performance deteriorated when digital keypoints were affected by occlusion. Our findings extend our prior work [18]; here, we establish the robustness of the end-to-end method across a broad range of tasks among participants who varied in hand size, sex, and skin tone (see Table 1), we quantify the differences in joint kinematics that result from these two reconstruction methods, and we demonstrate that the end-to-end approach achieves superior performance as evaluated by a computer vision metric.

Despite these advances, this work has several limitations. Most importantly, our quantitative error metric was not based on ground-truth 3D measurements, such as marker-based motion capture. Instead, reconstruction performance was quantitatively compared to the 2D digital keypoints with the highest likelihood scores available across all 8 camera views. Similarly, differences in reconstructed joint angles between methods were not quantitatively evaluated relative to an external reference standard. However, both reconstruction methods used the same video data, 2D digital keypoints, camera calibration matrices, and biomechanical model, providing context to their relative performance. Additionally, the participant cohort allowed the proposed framework to be evaluated across several sources of variability known to influence vision-based motion tracking methods.

The relative robustness of the end-to-end method suggests that it is more effective in utilizing the available 2D digital keypoint information. This capitalizes on recent advances in pose estimation algorithms, such as RTMPose [3], where increasingly accurate and diverse training datasets (Freihand [41], COCO-Wholebody [42], Hand10k [43], RHD2D [44], and COCO-Halpe [45]) improve keypoint localization and generalizability across a wide range of conditions. Because the end-to-end method is independent of the underlying pose estimator and dataset, future improvements in keypoint detection can be directly incorporated into the reconstruction pipeline to further improve reconstruction performance and reduce landmark localization errors. Importantly, previous studies have shown that small landmark localization and model scaling errors can propagate through the kinematic chain, leading to amplified errors in distal segments [46], [47], [48]. The reduced distal joint angle discrepancies observed with the end-to-end method suggest that jointly optimizing model pose and scaling helps mitigate this error propagation. Looking forward, emerging methods that reconstruct anatomically detailed hand surfaces rather than sparse keypoints alone [49], [50], may further improve markerless hand tracking by providing richer spatial information about the hand shape and articulation. Integrating these developments with biomechanics-aware end-to-end reconstruction methods can enhance the accuracy of dexterous hand and object interaction tasks. This work lays the foundation for more anatomically grounded motion capture systems that can be deployed in clinical and rehabilitation settings. By elimination manual post-processing while improving reconstruction accuracy, the end-to-end pipeline enables scalable biomechanical assessment of hand function.

## V. Conclusion

This study demonstrates the improvements of a single-stage, end-to-end optimization pipeline compared to a common two-stage reconstruction technique for the hand. Integration of the biomechanical constraints in the end-to-end approach enables accurate reconstruction of 3D hand trajectories from multi-view markerless motion capture. These findings establish a foundation for anatomically grounded approaches that can extend to bimanual models, diverse populations, and dexterous tasks. Ultimately, this work highlights the potential of video-based motion capture to transform clinical assessment, rehabilitation monitoring, and assistive technology development.

## VI. Acknowledgment

Research reported in this publication was supported by the National Institute of Neurological Disorders and Stroke of the National Institutes of Health under Award Number R01NS131953. The content is solely the responsibility of the authors and does not necessarily represent the official views of the National Institute of Health. Research reported in this publication was supported by the Eunice Kennedy Shriver National Institute Of Child Health & Human Development of the National Institutes of Health under Award Number T32HD007418.

AI Use Statement: The authors used Microsoft CoPilot to assist verbalizing the statistical analysis results, language editing, and drafting revisions. All outputs were reviewed and verified by the authors, who take full responsibility for the final manuscript.

Disclosure Statement: The authors report there are no competing interests to declare. Author Contributions Statement: P.F. designed the experimental study, integrated hand-specific biomechanical model into the MMC system, developed the hand specific tracking and data analysis scripts, and drafted the manuscript. J.P., R.J.C., and K.S. contributed to experimental data collection, initial data processing, and development of tools for the differentiable MMC system. A.S. provided the consultation and contributed to translation of the biomechanical model from OpenSim to Mujoco. L.E.M. and W.M.M. supervised the project and critically revised the manuscript. W.M.M. also contributed to study conceptualization and project oversight. All authors reviewed and approved the final manuscript and agreed to be accountable for all aspects of the work.